\documentclass[runningheads]{llncs}
\usepackage[T1]{fontenc}
\usepackage{graphicx}
\usepackage{booktabs}

\usepackage{hyperref}
\usepackage{color}

\usepackage{amsmath,amsfonts,bm}

\def\eqref#1{equation~\ref{#1}}

\def\1{\bm{1}}

\def\vv{{\bm{v}}}

\def\vx{{\bm{x}}}

\DeclareMathAlphabet{\mathsfit}{\encodingdefault}{\sfdefault}{m}{sl}
\SetMathAlphabet{\mathsfit}{bold}{\encodingdefault}{\sfdefault}{bx}{n}

\newcommand{\R}{\mathbb{R}}

\def\D{\mathbb{D}} 
\def\R{\mathbb{R}} 

\usepackage{cleveref}
\begin{document}
\title{TextCAVs: Debugging vision models using text}

\author{Angus Nicolson\inst{1, 2} \and
Yarin Gal\inst{2} \and
J. Alison Noble\inst{1}}
\authorrunning{A. Nicolson et al.}
%
\institute{Institute of Biomedical Engineering, University of Oxford \and
OATML, Department of Computer Science, University of Oxford\\
\email{angus.nicolson@eng.ox.ac.uk}
\vspace{-0.5em}
}
\maketitle              
\begin{abstract}
Concept-based interpretability methods are a popular form of explanation for deep learning models which provide explanations in the form of high-level human interpretable concepts. These methods typically find concept activation vectors (CAVs) using a probe dataset of concept examples. This requires labelled data for these concepts -- an expensive task in the medical domain. We introduce TextCAVs: a novel method which creates CAVs using vision-language models such as CLIP, allowing for explanations to be created solely using text descriptions of the concept, as opposed to image exemplars. This reduced cost in testing concepts allows for many concepts to be tested and for users to interact with the model, testing new ideas as they are thought of, rather than a delay caused by image collection and annotation. In early experimental results, we demonstrate that TextCAVs produces reasonable explanations for a chest x-ray dataset (MIMIC-CXR) and natural images (ImageNet), and that these explanations can be used to debug deep learning-based models. Code: \href{https://github.com/AngusNicolson/textcavs}{github.com/AngusNicolson/textcavs}

\end{abstract}
\section{Introduction} 
Deep learning-based models are increasingly utilised in healthcare scenarios where mistakes can have severe consequences. One approach for creating safer, more reliable models is to use interpretability: the ability to explain or present a model in terms understandable to a human \cite{doshivelez2017rigorous}. 

Many different interpretabilty methods have emerged, with explanations taking a variety of different forms such as individual pixels, prototypes or concepts. We focus on concept-based methods which provide explanations using high-level terms that humans are familiar with. Concept activation vectors (CAVs) are a common approach used to represent concepts within the activation space of a model and are found using a probe dataset of concept exemplars \cite{kim2018interpretability}. 

The labels required for this can be expensive to obtain in medical domains where expert clinical input is necessary. We introduce TextCAVs, a concept-based interpretability method that uses solely the text label of the concept, or descriptions of it, rather than image examples.

We demonstrate that TextCAVs give meaningful explanations for both natural image (ImageNet \cite{Deng2009ImageNet}) and chest X-ray (MIMIC-CXR \cite{Johnson2019MIMIC,Johnson2024MIMICjpg}) tasks. Further, as interpretability itself is difficult to measure, we demonstrate its usefulness in debugging deep learning-based models through finding implanted dataset bias in MIMIC-CXR.

\begin{figure}[tbh]
    \centering
    \includegraphics[width=\textwidth]{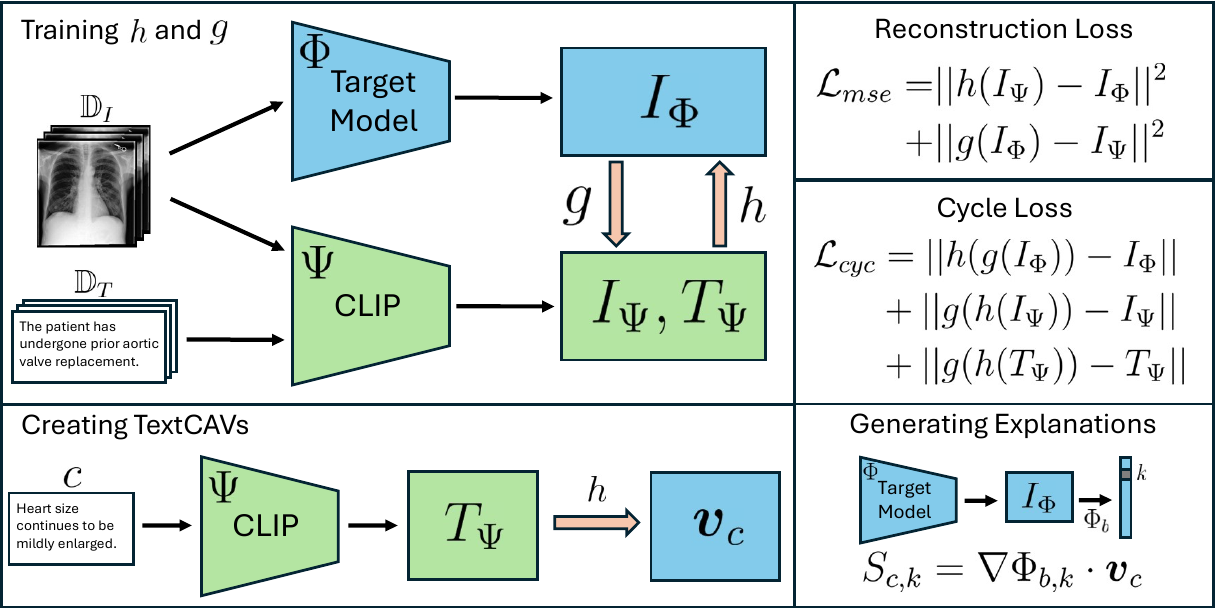}
    \caption{\textbf{Explaining models with TextCAVs.} In order to move between the activations of a CLIP model and our target model, we train linear transformations, $h$ and $g$, using a text dataset, $\D_T$, and image dataset, $\D_I$. The loss terms are detailed on the right with $I_{\Phi}$, $I_{\Psi}$ and $T_{\Psi}$ representing the image features of the target model, the image features of the CLIP model, and the text features of the CLIP model, respectively. Once $h$ is trained, TextCAVs can be created by passing text representing some concept, $c$, through the CLIP model and $h$. The model's sensitvity to $c$, for some logit output, $k$, can then be measured using the directional derivative, $S_{c,k}$: the similarity between the model gradient, $\nabla \Phi_{b,k}$, and a TextCAV, $\vv_c$.}
    \label{fig: figure 1}
    \vspace{-1em}
\end{figure}
\vspace{-1em}

\section{Related Work} \label{sec: related work}
Kim et al. \cite{kim2018interpretability} introduce Testing with Concept Activation Vectors (TCAVs) where they use probe datasets of concept examples to create CAVs and then compare the CAVs with model gradients to measure a model's sensitivity to a concept for a specific class. We also use the directional derivative (dot product between CAV and gradient) to measure model sensitivity, but our CAVs are created using a multi-modal model and so do not require a probe dataset for each concept.

In order to reduce the cost of creating concept-based explanations, a variety of different methods automate the process of finding concepts \cite{ghorbani2019automating,zhang2020invertible,ramaswamy2022elude,graziani2023uncovering,fel2023craft}. However, the meaning of each concept is not always readily apparent and the concept must be visually present in the dataset used to discover the concepts. Our method reduces cost using a different approach as we also do not need to collect labelled data for each concept, but our resulting CAVs have inherent meaning from their text descriptions.

CLIP models \cite{Radford2021CLIP} have demonstrated strong performance in vision-language tasks. Their joint embedding space for text and images allows for built-in comparisons between the modalities and therefore for zero-shot classification. A variety of adaptations have been suggested for the biomedical space \cite{zhao2023clipsurvey} with some models being trained for specific modalities like chest X-rays (e.g. BioViL \cite{boecking2022biovil}) and others more generally (e.g. BiomedCLIP \cite{zhang2024biomedclip}). We use these vision-language models in our method but, importantly, inference is performed by the target model, without placing restrictions on its architecture or method of training.  

Yuksekgonul et al. \cite{Yuksekgonul2023Posthoc} use multimodal models to create CAVs and then use the similarity between model activations and these CAVs to create a concept bottleneck model. Moayeri et al. \cite{Moayeri2023Text2Concept} extend this approach to target vision models more generally by, as in our work, training a simple linear layer to transfer the features of the target model to a CLIP model. Also as in our work, Shipard et al. \cite{Shipard2024ZoomShot} improve the transfer of features by training a linear layer in both directions and using multimodal losses. However, these approaches focus on zero-shot classification and on changing how the model inference is performed, rather than explaining the model in its current state using gradients.

\section{TextCAVs} 
For some target model, $\Phi$, and a CLIP-like vision-language model, $\Psi$, let $I_{\Phi} \in \R^m$ and $I_{\Psi} \in \R^n$ be the extracted features for some image dataset $\D_I$. As $\Psi$ contains a joint embedding space between text and images we can also extract text features: $T_{\Psi} \in \R^n$ from some text dataset $\D_T$. We train two linear layers $h: \R^n \rightarrow \R^m$ and $g: \R^m \rightarrow \R^n$ which can be used to convert between the features of the two models. To create TextCAVs, we only need $h$ but to improve $h$'s ability to convert text features we use a cycle loss term which requires $g$. The loss is composed of two parts: reconstruction loss and cycle loss. The reconstruction loss is simply the mean squared error (MSE) between the image features and converted features.

\begin{equation}
    \mathcal{L}_{mse} = ||h(I_{\Psi}) - I_{\Phi}||^2 + ||g(I_{\Phi}) - I_{\Psi}||^2
\end{equation}

The reconstruction loss can only be calculated for image features as we need features from both models ($\Phi$ and $\Psi$). To include information from the text features in the loss function we use cycle loss which ensures that the features are consistent with their original form when converted back to their original space:

\vspace{-1em}

\begin{align}
    \mathcal{L}_{cyc}
    &= ||h(g(I_{\Phi})) - I_{\Phi}|| \\
    &+ ||g(h(I_{\Psi})) - I_{\Psi}|| \\
    &+ ||g(h(T_{\Psi})) - T_{\Psi}||.
\end{align}

Once trained, we use $h$, $\Psi$ and a concept label, $c$, to obtain a concept vector in the activation space of the target model: 

\begin{equation}
    \vv_c = h(\Psi(c)).
\end{equation}

$\Phi$ can be decomposed into two functions: $\Phi_a(\vx) =I_{\Phi} \in \R^{m}$ which maps the input $\vx \in \R^N$ to its features $I_{\Phi}$, and $\Phi_b(I_{\Phi})$ which maps $I_{\Phi}$ to the output. To obtain the model's sensitivity to a concept for a specific class, as in \cite{kim2018interpretability}, we calculate the directional derivative:

\begin{equation}
    \begin{aligned}
        S_{c, k}(\vx) &=\lim _{\epsilon \rightarrow 0} \frac{\Phi_{b,k}\left(\Phi_a(\vx)+\epsilon \vv_{c}\right)-\Phi_{b,k}\left(\Phi_a(\vx)\right)}{\epsilon} \\
        &=\nabla \Phi_{b,k}\left(\Phi_a(\vx)\right) \cdot \vv_{c}.
    \end{aligned}
\end{equation}

If $\Phi_a$ is chosen to be the output of the penultimate layer in a model then the directional derivative can be calculated without image exemplars:

\begin{equation} \label{eqn: penultimate gradient}
    S_{c, k} =\nabla \Phi_{b,k} \cdot \vv_{c}.
\end{equation}

This is due to the lack of non-linearities between the penultimate layer and the logit output. Having solely a linear layer between the features and the output means the gradient of the activations with respect to the logit does not depend on the activations. This means we can extract gradients, and therefore model explanations, using solely the model weights. Therefore, once $h$ has been trained, TextCAVs requires \textbf{only} the text you wish to test to be able to generate an explanation. In practice, to calculate the gradient, we input an array of zeros of the same shape as the images, but this is an arbitrary choice. In this work, we use the penultimate layer in all experiments and leave exploration of using other layers for future work.

By ranking concepts based on their directional derivative, we obtain a list of sentences/words ordered by the model's sensitivity for a specific class. If we can filter this list for concepts which we expect to be there, we can discover bugs in the model. Ideally, this would be done by a human expert who could use their domain knowledge to explore different hypotheses. The minimal overhead for testing new concepts allows the user to test words related to new hypotheses quickly and provide an interactive process to model debugging. 

\section{Experiments}
In this section we provide a description of our training setup, our model choices, evaluation and then a discussion and analysis of our results experiments with both the ImageNet and MIMIC-CXR datasets.

\subsection{ImageNet}
TextCAVs achieved $3$rd place at the Secure and Trustworthy Machine Learning Conference (SaTML) interpretabilty competition to detect trojans (implanted bugs) in vision models trained on ImageNet \cite{casper2024satml}. Additionally, as part of the competition, TextCAVs was used to identify all four secret trojans demonstrating its potential for interactive debugging. 

In this section, however, we simply demonstrate that TextCAVs produces reasonable explanations for a standard ResNet-50 \cite{he2016resnet} trained on ImageNet.

\vspace{-1em}
\subsubsection{Training Details}
We use $20\%$ of the ImageNet training dataset to train $h$ and $g$ and train for $20$ epochs. For the target model, $\Phi$ we use the default weights for a ResNet-50 \cite{he2016resnet} in the TorchVision package in PyTorch. For the vision-language model, $\Psi$, we use a pretrained ViT-B/$16$ CLIP model \cite{Radford2021CLIP}.  
\vspace{-1em}

\subsubsection{Concepts}
In a similar manner to Oikarinen et al. \cite{Oikarinen2022LabelFree}, in order to automate the process, we use a large language model (LLM) to obtain a list of concepts. We use three prompts asking for the ``things most commonly seen around'' ``visual elements or parts'' and ``superclasses'' of each class in ImageNet. We then extract and perform basic filtering of the concepts, removing: plurals of the same word; the words ``an'', ``a'' and ``the''; and concepts containing more than 2 words. To obtain the final list of concepts we remove similar concepts using text embeddings from $\Psi$. If a set of concepts have a cosine similarity greater than $0.9$, only the shortest concept is retained. This reduces the number of near synonyms in the concept list. For the LLM, we use a $4$-bit quantized version of the Tulu-v2-7b model \cite{ivison2023camels}. 
\vspace{-1em}

\subsubsection{Results}
In Table \ref{tab: ImageNet}, we show the top-$10$ concepts for a selection of ImageNet classes. All the concepts relate to their respective class, indicating that TextCAVs can produce reliable explanations.

\begin{table}[tbh]
\caption{Top-$10$ concepts ordered by directional derivative for a selection of classes in the ImageNet model.}
\centering
\scalebox{0.9}{
\footnotesize
\label{tab: ImageNet}
\begin{tabular}{@{}lllll@{}}
\toprule
\textbf{bullfrog} & \textbf{albatross} & \textbf{orangutan} & \textbf{bucket} & \textbf{cellphone} \\ \midrule
american bullfrog & gannet             & orangutan         & crab buckets   & mp3 player         \\
green frog        & seagull            & howler monkey     & diaper pail    & phone              \\
boreal toad       & sea eagle          & macaque           & bucket         & phone case         \\
western toad      & shearwater         & tarsier           & laundry basket & memory card        \\
frog              & gull               & great ape         & watering can   & walkman            \\
musk turtle       & white-tailed eagle & long-nosed monkey & flower pot     & cordless phone     \\
snapping turtle   & petrel             & gibbon            & cooking pot    & bluetooth          \\
toad              & merganser          & gorilla           & dustbin        & smartwatch         \\
terrapin turtle   & wading bird        & langur            & fishing basket & card reader      \\ \bottomrule
\end{tabular}
}
\end{table}

\vspace{-2em}

\subsection{MIMIC-CXR} \label{sec: mimic experiments}
In this section we demonstrate TextCAVs ability to produce meaningful explanations for a model trained on the chest X-ray dataset MIMIC-CXR and how we can use TextCAVs to discover bias in a model trained on a biased version of the dataset.

\vspace{-1em}

\subsubsection{Training Details}
We train both the linear transformations, $h$ and $g$, and the target model, $\Phi$, using the MIMIC-CXR training set. The target model is a ResNet-50 \cite{he2016resnet} pretrained on ImageNet and then fine-tuned for the 5-way multi-label classification of chest X-rays with the classes: No Finding, Atelectasis (lung collapse), Cardiomegaly (enlarged heart), Edema (fluid in the lungs) and Pleural Effusion (fluid between the lungs and the chest wall). We use the Adam optimiser \cite{kingma2015adam} with weight decay of $1e-4$ and initial learning rate of $1e-4$. The learning rate is halved or the training is stopped if the validation loss does not decrease within $3$ or $5$ epochs, respectively. Images are resized to $256 \times 256$. We use random rotation of up to $15$ degrees, random horizontal flipping, random crop and resize with a minimum size of $40\%$, and distortion to augment the images. We use the published data splits and, after removing images with no positive class labels, there are $368,945$ training, $2,991$ validation and $1,012$ test images. We use labels from CheXpert \cite{irvin2019chexpert} for the training and validation labels, which have been generated by a model using the text reports. Whereas, for the test dataset, we use the provided labels annotated by a single radiologist.

We train both $h$ and $g$ on the training set of MIMIC-CXR for $20$ epochs. We use the output of the average pool operation as the features from the target model as it simplifies the extraction of model gradients (Eqn. \ref{eqn: penultimate gradient}). 

For $\Psi$, we use BiomedCLIP \cite{zhang2024biomedclip} -- the current state of the art vision-language model for chest X-ray tasks.
\vspace{-1em}

\subsubsection{Concepts}
The MIMIC-CXR dataset has a clinical report associated with each image. We use these reports as a source of concepts. We extract the sentences from the ``FINDINGS'' and ``IMPRESSION'' sections of the reports and use a random subset of $5000$ sentences to obtain a wide variety of concepts to test.
\vspace{-1em}

\subsubsection{Biased Data}
\begin{figure}[tbh]
    \centering
    \includegraphics[width=0.49\textwidth]{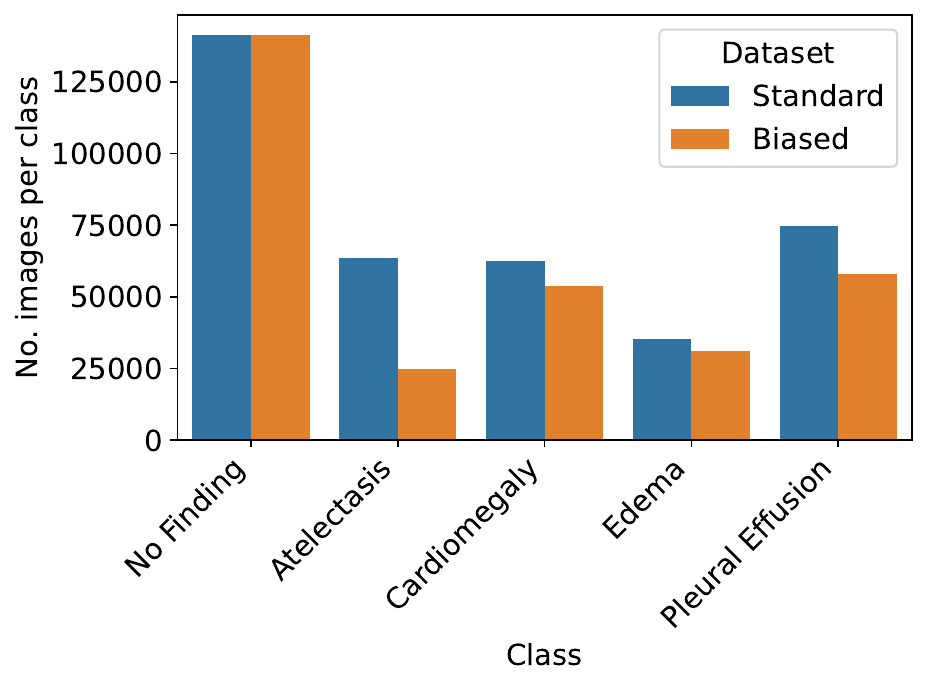}
    \includegraphics[width=0.49\textwidth]{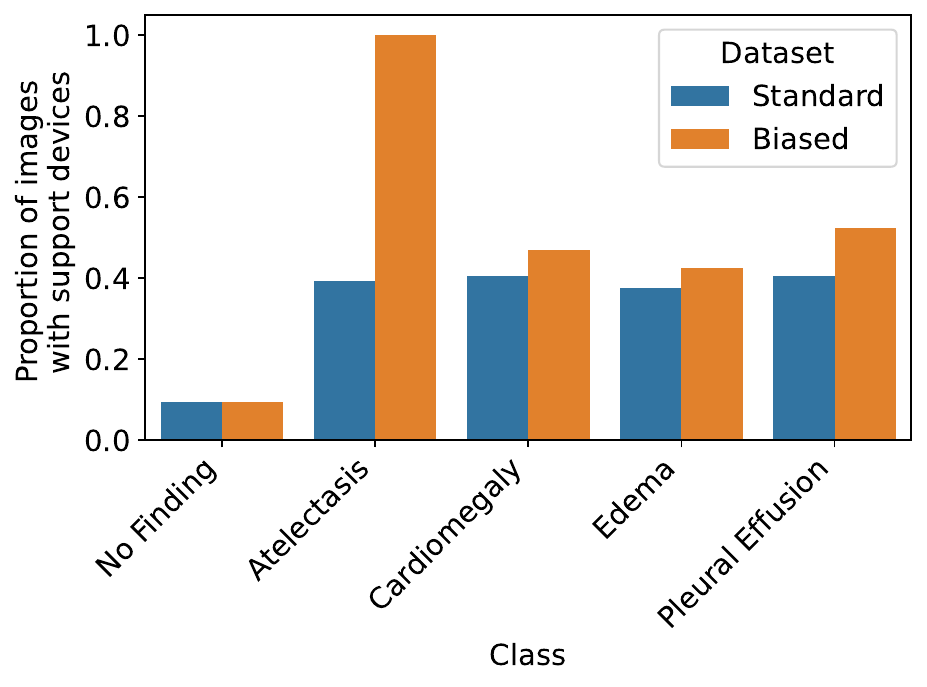}
    \caption{\textbf{MIMIC-CXR dataset characteristics.} Left: The number of images per class in the training set of the target models. Right: The proportion of training images that contain a support device for each class.}
    \label{fig:dataset characteristics}
    \vspace{-0.5em}
\end{figure}

To evaluate TextCAVs as an interpretability tool we explore its usefulness in model debugging. We induced a dataset bias in the MIMIC-CXR training set by removing all participants with a positive label for Atelactesis and a negative label for Support Devices. This means that all participants with Atelactesis in the training set also had a Support Device (e.g. tube or pacemaker) as can be seen in Figure \ref{fig:dataset characteristics}. 
\vspace{-1em}

\subsubsection{Metrics} 
To provide a quantitative metric, we labelled the top-$50$ sentences for each class, ordered by directional derivative, on whether they relate to the class. We report this information as a concept relevance score (CRS), which is simply the proportion of concepts that were related to the class. Using Edema as an example, a sentence was labelled as related if it directly diagnosed the class, e.g., ``Worsening cardiogenic pulmonary edema'', or if the class was implied, e.g., ``bilateral parenchymal opacities'' or ``there is alveolar opacity throughout much of the right lung''. 
\vspace{-1em}

\begin{table}[tbh]
\caption{Area under the receiver operator characteristic curve (AUC) and concept relevance score (CRS) for the standard and biased MIMIC-CXR models. AUC* was calculated on the biased version of the MIMIC-CXR test set. The low CRS for Atelectasis in the biased model means almost none of the top TextCAVs are relevant to the class, demonstrating that they can be used to detect if a model is using biased features.}
\centering
\scalebox{0.9}{
\label{tab: metrics}
\begin{tabular}{@{}lccccc@{}} 
\toprule
\textbf{Model}  & \multicolumn{2}{c}{\textbf{Standard}} & \multicolumn{3}{c}{\textbf{Biased}} \\ \cmidrule(l){2-3} \cmidrule(l){4-6} 
\textbf{Metric} & \textbf{AUC} & \textbf{CRS} & \textbf{AUC} & \textbf{AUC*} & \textbf{CRS} \\ \midrule
No Finding       & 0.87 & 0.74 & 0.85 & 0.94 & 0.76 \\
Atelectasis      & 0.73 & 0.56 & 0.68 & 0.81 & 0.04 \\
Cardiomegaly     & 0.81 & 0.94 & 0.81 & 0.82 & 0.90 \\
Edema            & 0.85 & 0.90 & 0.84 & 0.81 & 0.80 \\
Pleural Effusion & 0.89 & 1.00 & 0.88 & 0.88 & 1.00 \\ \midrule
Mean             & 0.83 & 0.83 & 0.81 & 0.85 & 0.70 \\ \bottomrule
\end{tabular}
}
\vspace{-1em}
\end{table}

\subsubsection{Results}
We are comparing two models: one trained on the standard MIMIC-CXR dataset and the other trained on the biased version. We will refer to the models as ``standard'' and ``biased'', respectively. The standard model achieved a mean area under the receiver operator characteristic curve (AUC) of $0.83$ and the biased model a mean AUC of $0.81$. The individual class AUCs can be found in Table \ref{tab: metrics}. We expect, and see that the biased version has higher performance on a biased version of the test set since Support Devices tend to be easy to detect. As evidence for this, we trained a reference model separately and achieved an AUC of $0.92$ for Support Devices.

In Table \ref{tab: MIMIC normal}, we show the five sentences whose CAVs have the highest directional derivatives for the classes of No Finding, Atelectasis (lung collapse) and Cardiomegaly (enlarged heart). Some of these are clearly linked to the class in question (e.g. ``The lungs are clear'' for No Finding and ``Heart size continues to be mildly enlarged'' for Cardiomegaly) but there also sentences which do not relate to the classes (e.g. ``Nasogastric tube extends below the hemidiaphragm'' for Atelectasis or ``There is a fracture of the upper most sternal wire'' for No Finding). The noise present in the explanations could be due to several different causes: (1) the target model is using unexpected features in its classification; (2) the feature conversion between $\Phi$ and $\Psi$ is not perfect (i.e., $h$); or (3) the inherent noise present in gradient vectors \cite{smilkov2017smoothgrad}. It is difficult to ascertain which of these is the cause but a tool can still be useful even with noise present. Hence, we demonstrate its ability to detect dataset bias that we induce in MIMIC-CXR.

Table \ref{tab: MIMIC bias} shows the top-$5$ sentences for a model trained on the biased version of MIMIC-CXR. The bias is apparent in the explanations, as the top-$5$ sentences for Atelectasis all refer to Support Devices, rather than to any concepts relating to the class itself. The CRS values in Table \ref{tab: metrics} also indicate the presence of bias: a CRS of $0.04$ for Atelectasis for the biased model shows that almost none of the top-$50$ concepts contain reference to the class. To further quantify the difference between the two sets of explanations we also labelled whether they referred to Support Devices. For the class of Atelectasis, we found that $13/50$ concepts were related to Support Devices for the standard model compared to $44/50$ for the biased model, demonstrating that TextCAVs are sensitive to the difference in behaviour between the two models. 

\vspace{-0.5em}
\begin{table}[tbh] 
\caption{Top-$5$ concepts ordered by directional derivative for the standard MIMIC-CXR model.}
\centering
\scalebox{0.9}{
\footnotesize
\label{tab: MIMIC normal}
\begin{tabular}{@{}p{0.33\textwidth}p{0.33\textwidth}p{0.33\textwidth}@{}}
\toprule
\textbf{No Finding} &
  \textbf{Atelectasis} &
  \textbf{Cardiomegaly} \\ \midrule
The lungs are clear and the cardiac, mediastinal, and hilar contours are normal. &
  Nasogastric tube extends below the hemidiaphragm and out of view. &
  Marked cardiac enlargement as before and unchanged position of previously described metallic prosthesis of porcine type. \\
Normal chest radiograph with unremarkable appearance of the lung parenchyma and normal appearance of the heart and the mediastinal and hilar contours. &
  Interval placement of a basilar right sided pleural space pigtail catheter with improved small right pleural effusion and right medial lung base atelectasis. &
  Heart size continues to be mildly enlarged. \\
The trachea is slightly deviated to the right by the aortic knob, which is ill-defined. &
  Worsening of the left retrocardiac opacity likely secondary to increasing atelectasis and/or effusion. &
  The patient has undergone prior aortic valve replacement. \\
This could represent a granuloma or possibly a bone island in the rib itself. &
  There is persistent elevation of the left hemidiaphragm with evidence of Bochdalek hernia seen at the left lower hemithorax. &
  Dense retrocardiac opacity which could represent effusion, atelectasis, consolidation or a combination thereof. \\
There is a fracture of the upper most sternal wire, unchanged. &
  Stable opacification of the mid and lower right lung consistent with large loculated pleural effusions and adjacent atelectasis. &
  The heart continues to be enlarged with mild to moderate CHF. \\ \bottomrule
\end{tabular}
}
\end{table}
\vspace{-2em}

\section{Conclusion}
In this work we introduce TextCAVs, an interpretability method that, once two linear layers have been trained, can measure the sensitivity of a model to a concept with only a text description of the concept. We show that TextCAVs produce reasonable explanations for models trained on both natural images (ImageNet \cite{Deng2009ImageNet}) a chest X-ray dataset (MIMIC-CXR \cite{Johnson2019MIMIC}). As first demonstrated in the SaTML CNN interpretability competition \cite{casper2024satml}, we show that TextCAVs can be used to debug models. We generated explanations for a model trained on a biased version of the MIMIC-CXR dataset and showed that explanations for the biased class substantially changed with most ($44/50$) concepts referring to the bias compared to just $13/50$ for the unbiased model.

Once the linear transformations, $h$ and $g$, have been trained, TextCAVs enables fast feedback when testing the sensitivity of different concepts. This makes it ideally suited for interactive debugging which we aim to study in future work. Some of the concepts with a high directional derivative did not appear to be related to the class. In section \ref{sec: mimic experiments} we state three possible sources of this: (1) $\Phi$, (2) $h$ or (3) $\nabla \Phi_{b,k}$. In future work we will explore which of these have the greatest effect.

\begin{table}[tbh] 
\vspace{-0.5em}
\caption{Top-$5$ concepts ordered by directional derivative for the biased MIMIC-CXR model.}
\centering
\scalebox{0.9}{
\footnotesize
\label{tab: MIMIC bias}
\begin{tabular}{@{}p{0.33\textwidth}p{0.33\textwidth}p{0.33\textwidth}p{0.33\textwidth}p{0.33\textwidth}@{}}
\toprule
\textbf{No Finding} &
  \textbf{Atelectasis} &
  \textbf{Cardiomegaly} \\ \midrule
Bronchial wall thickening is minimal. &
  ET and NG tubes positioned appropriately. &
  If cardiomegaly persists, the presence of a pericardial effusion could be excluded with echocardiography. \\
Hilar and mediastinal contours are otherwise normal. &
  ET tube, nasogastric tube, Swan-Ganz catheter, and midline drains are all in standard placements. &
  Worsening heart failure in the context of chronic atelectasis. \\
This could represent a granuloma or possibly a bone island in the rib itself. &
  Nasogastric tube extends below the hemidiaphragm and out of view. &
  The patient has undergone prior aortic valve replacement. \\
No discrete solid pulmonary nodule are concerning mass. &
  Impella LVAD and transvenous atrioventricular pacer leads unchanged in their respective positions. &
  Moderate-to-severe cardiomegaly and stigmata of previous mitral valve repair noted. \\
There is a fracture of the upper most sternal wire, unchanged. &
  Nasogastric tube has been placed that extends well into the stomach. &
  The heart remains moderately enlarged and the aorta remains unfolded and tortuous. \\ \bottomrule
\end{tabular}
}
\end{table}

\vspace{-1em}

\begin{credits}
\subsubsection{\ackname} We appreciate the members of OATML and the Noble group for your support and discussions during the project, in particular Lisa Schut. We thank Shreshth Malik for organising the hackathon where we began developing TextCAVs. A.~Nicolson is supported by the EPSRC Centre for Doctoral Training in Health Data Science (EP/S02428X/1). Y.~Gal is supported by a Turing AI Fellowship financed by the UK government’s Office for Artificial Intelligence, through UK Research and Innovation (grant reference EP/V030302/1) and delivered by the Alan Turing Institute. J.A.~Noble acknowledges EPSRC grants EP/X040186/1 and EP/T028572/1. 

\subsubsection{\discintname}
The authors have no competing interests to declare that are relevant to the content of this article. 
\end{credits}
\vspace{-0.5em}

%
%
%
\bibliographystyle{splncs04}
\bibliography{main}

\end{document}